\documentclass{isprs} 
\usepackage{setspace}
\usepackage{geometry} 
\usepackage{epstopdf}
\usepackage[labelsep=period]{caption}  
\usepackage[british]{babel} 
\usepackage[hang]{footmisc}

\usepackage{enumitem}
\usepackage[authoryear,round]{natbib}

\usepackage{acro}
\DeclareAcronym{uq}{
    short=UQ,
    long=uncertainty quantification
}
\DeclareAcronym{de}{
    short=DE,
    long=Deep Ensembles
}
\DeclareAcronym{mcdo}{
    short=MC Dropout,
    long=Monte-Carlo Dropout
}
\DeclareAcronym{nll}{
    short=NLL,
    long=negative log-likelihood
}
\DeclareAcronym{bnn}{
    short=BNNs,
    long=Bayesian neural networks
}
\DeclareAcronym{bop}{
    short=BOP,
    long=Benchmark for 6D Object Pose Estimation
}
\DeclareAcronym{pnp}{
    short=P$n$P,
    long=Perspective-$n$-Point
}
\DeclareAcronym{vat}{
    short=VAT,
    long=virtual adversarial training
}
\DeclareAcronym{uscore}{
    short=UCS,
    long=uncertainty calibration score
}
\DeclareAcronym{cnn}{
    short=CNN,
    long=convolutional neural network
}
\DeclareAcronym{ece}{
    short=ECE,
    long=expected calibration error
}

\usepackage{graphicx}
\usepackage[subrefformat=parens,labelformat=parens]{subfig}
\usepackage{amsmath}
\usepackage{amssymb}
\usepackage{multirow}
\usepackage{floatrow}
\DeclareMarginSet{hangboth}{\setfloatmargins*{\hskip-20mm}{\hskip-20mm}}
\usepackage[pagebackref,breaklinks]{hyperref}

\newcommand{\Unit}[1]{\thinspace #1}

\begin{document}

\title{Uncertainty Quantification with Deep Ensembles \\for 6D Object Pose Estimation}
\date{}

\author{Kira Wursthorn\thanks{Corresponding author}, Markus Hillemann, Markus Ulrich}

\address{Institute of Photogrammetry and Remote Sensing (IPF), Karlsruhe Institute of Technology (KIT), Germany\\
\{kira.wursthorn, markus.hillemann, markus.ulrich\}@kit.edu}


\commission{II/}{YY} 
\workinggroup{XX/YY} 
\icwg{}   

\abstract{
The estimation of 6D object poses is a fundamental task in many computer vision applications. Particularly, in high risk scenarios such as human-robot interaction, industrial inspection, and automation, reliable pose estimates are crucial. In the last years, increasingly accurate and robust deep-learning-based approaches for 6D object pose estimation have been proposed. Many top-performing methods are not end-to-end trainable but consist of multiple stages. In the context of deep uncertainty quantification, deep ensembles are considered as state of the art since they have been proven to produce well-calibrated and robust uncertainty estimates. However, deep ensembles can only be applied to methods that can be trained end-to-end. In this work, we propose a method to quantify the uncertainty of multi-stage 6D object pose estimation approaches with deep ensembles. For the implementation, we choose \mbox{SurfEmb} as representative, since it is one of the top-performing 6D object pose estimation approaches in the BOP Challenge 2022. We apply established metrics and concepts for deep uncertainty quantification to evaluate the results. Furthermore, we propose a novel uncertainty calibration score for regression tasks to quantify the quality of the estimated uncertainty.
}


\maketitle


\section{Introduction}\label{sec:introduction}
Determining the 6D pose of an object in its 3D environment, \mbox{i.e.}, its 3D orientation and 3D position, from a camera image or from depth sensor data is a fundamental task in computer vision. Applications like augmented reality~\citep{yongzhi2019ar}, vision-assisted robot manipulation~\citep{steger2018machine_vision,ulrich2024hand_eye_calibration}, bin picking~\cite{drost2017bin_picking}, and autonomous systems as self-driving cars~\citep{yurtsever2020survey} rely on accurate object poses estimated from RGB(-D) images. In the real world, complex scenes arise that lead to typical challenges for 6D object pose estimation: The scene might be cluttered with multiple instances of varying object categories, sometimes with more than one instance of a given object category and occluded instances. Objects might be symmetric or contain inter-object similarity, where one object is built from parts of other objects. In addition, object surfaces can be challenging if they are textureless or reflective, for example. 
The \ac{bop} and the associated BOP Challenge 2020~\citep{hodan2020bop_challenge} and 2022~\citep{sundermeyer2023bop} cover these challenges and allow a robust evaluation of state-of-the-art methods.
These methods incorporate deep learning at large, taking advantage of the ability of deep neural networks to learn complex patterns on sufficient amounts of data and use RGB images as well as depth information to compute the 6D object pose. 
Many of the top-performing approaches \citep{park2019pix2pose,labbe2020cosypose,haugaard2022surfemb,wang2021gdr_net} are composed of three major stages:
First, an off-the-shelf object detector locates the target object instance in the image of the scene. Second, a deep neural network predicts the 2D--3D correspondences, and third, a variant of the \ac{pnp} algorithm, often combined with RANSAC, provides the 6D pose of the detected object instances. Optionally, depth information is used for pose refinement.

In safety-critical applications like autonomous driving~\citep{mcallister2017autonomous_driving}
and demanding industrial applications like quality inspection and automation~\citep{heizmann2022industrial_inspection}
, the prediction of an object pose is often not sufficient to make informed decisions. 
Instead, the associated object pose uncertainty must also be taken into account.
For example, consider the task of a robot grasping a cup whose pose is estimated based on a RGB(-D) input image that does not show the handle of the cup. This leads to an ambiguous pose estimate. If the robot grasps the cup based on that pose estimate, the object or the robot might be damaged. In combination with a measure of pose uncertainty, this scenario can be identified and prevented by choosing a different camera angle or, in a bin picking application, choosing another object to grasp that has a lower uncertainty.
In deep learning, popular \ac{uq} methods include softmax probability ~\citep{hendrycks2017a}, Monte-Carlo Dropout~\citep{gal2016dropout_bayesian}, and Deep Ensembles~\citep{deep_ensembles}. While softmax predictions are only used for classification and segmentation tasks, Monte-Carlo Dropout and Deep Ensembles can be applied to regression tasks as well, and therefore are suited for object pose estimation.
Deep ensembles of random initialized networks perform best and are more robust under datashift, compared to dropout methods, post-hoc calibration by temperature scaling, and methods motivated by Bayesian inference \citep{ovadia2019dataschift_uq_comparison}.

The application of \ac{uq} methods to multi-stage approaches for 6D object pose estimation is not straightforward. These methods are usually designed for segmentation and classification tasks, which often are single-stage approaches in the sense that they are end-to-end trainable.
Since 6D object poses have one orientation component in $\mathrm{SO(3)}$ and one position component in $\mathbb{R}^3$, the object pose is often considered in a decoupled fashion, handling orientation and position separately. While it can be generally assumed that the object position in $\mathbb{R}^3$ is normally distributed, modelling the orientation distribution is more complex. 
Considering this, Deep Ensembles and \ac{uq} methods that draw samples from the posterior predictive distribution have the advantage that no assumptions concerning the underlying distributions have to be made.
Surprisingly, up to now, there is no work that uses a deep ensemble approach for \ac{uq} in object pose estimation. The most closely related method by \citet{shi2021fast_uq} uses two to three heterogeneous pretrained pose estimation models to estimate the pose disagreement. However, while this approach reduces the computational cost of training and inferring a large ensemble of models, this approach diverges from the deep ensemble methodology and does not produce uncertainty estimates. 

 In this work, we propose a method to quantify the uncertainty of multi-stage 6D object pose estimation approaches with the current state-of-the-art deep learning \ac{uq} method, namely deep ensembles. For the implementation, we choose \mbox{SurfEmb}~\citep{haugaard2022surfemb}, a top-performing 6D object pose estimation method. We evaluate the estimated pose results and their uncertainties using reliability diagrams and \ac{bop} metrics on the T-LESS~\citep{hodan2017tless} and YCB-V~\citep{xiang2018pose_cnn} benchmark datasets for object pose estimation. Furthermore, we introduce a novel metric for the evaluation of uncertainty estimates in regression tasks in general.

\section{Related work}
As \ac{uq} in deep learning and explainable AI gain more and more interest, works on the reliability of both network predictions and estimated uncertainties have increased in the recent years. In Section \ref{subsec:uq}, an overview of popular \ac{uq} approaches in deep learning in general is given. In Section \ref{subsec:uqpose}, works on object pose uncertainties and pose distributions are described.

\subsection{\ac{uq} in Deep Learning}\label{subsec:uq}
\ac{uq} in deep learning often distinguishes between different types of uncertainties depending on their source. The predictive uncertainty is often split into aleatoric and epistemic uncertainty. Aleatoric uncertainty captures the uncertainty that is inherent in the input data, \mbox{e.g.} noise in an image, while epistemic uncertainty refers to a lack of knowledge, \mbox{i.e.} the uncertainty of the network parameters \citep{kendall2017cv_uncertainties}.

Deep-learning-based approaches for object pose estimation integrate large deep neural networks in their pipe\-lines in most cases. Consequently, these networks consist of a large amount of parameters and non-linearities that make the computation of the exact posterior probability distributions of the network's predicted outputs generally intractable \citep{blundell2015bayes_by_backprop,loquercio2020uncertainty_framework_dronet}. As a conse\-quence, approximation approaches are used for \ac{uq}.
Approaches based on Bayesian inference transform common deterministic networks into stochastic ones by placing probability distributions over either the activations and/or the weight parameters \citep{jospin2020hands_on_baysian_networks}, leading to \ac{bnn}. While \ac{bnn} have a mathematically sound foundation, the high parameter counts of deep neural networks make a direct solution impossible.

Bayes by Backprop~\citep{blundell2015bayes_by_backprop} is one work proposing variational inference to learn the parameters of approximate distributions over the weights. At inference time, weights are sampled from the learned distributions resulting in an ensemble of networks that is used to sample the posterior distribution of the predictions.
Because \ac{bnn} come with a high computational cost, Monte-Carlo Dropout~\citep{gal2016dropout_bayesian} was proposed where dropout regularization~\citep{Srivastava2014dropout} at inference time approximates a stochastic Gaussian process. Then, the posterior predictive distribution is samp\-led from multiple forward passes through networks with varying dropout masks requiring additional runtime at inference time. Furthermore, Monte-Carlo Dropout capture only the epistemic uncertainty and, therefore, has been combined with probabilistic networks \citep{kendall2017cv_uncertainties} and assumed density filtering~\citep{gast2018adf,loquercio2020uncertainty_framework_dronet} for predictive \ac{uq}.
Another drawback of Monte-Carlo Dropout is that the estimated uncertainties need to be calibrated \citep{gal2016dropout_bayesian}. In contrast, Deep Ensembles do not require a calibration \citep{deep_ensembles}. They present a robust way to estimate predictive uncertainty in computer vision tasks such as classification, semantic segmentation, and depth estimation, and are considered state-of-the-art in deep learning \ac{uq} \citep{ovadia2019dataschift_uq_comparison,gustafsson2020eval_bayesian_methods,wursthorn2022}. Ensemble distillation can be used to overcome the high memory costs of ensembles while achieving comparable uncertainty results \citep{landgraf2023dudes}.
Recently, \citep{mukhoti2023ddu} proposed a deterministic \ac{uq} approach that provides similar results to deep ensembles, even on out-of-dis\-tribution examples.

\subsection{\ac{uq} for Object Pose Estimation}\label{subsec:uqpose}
Despite the importance of reliable pose estimates, there are few works that focus on \ac{uq} in the context of 6D object pose estimation \citep{thalhammer2023challenges_object_pose_estimation}. Often, \ac{uq} for object pose estimation is referred to as the estimation of a pose distribution. 
Many works use Bingham distributions to model the orientation distribution \citep{gilitschenski2020binghamloss,okorn2020learning_orientation_distributions,deng2022bingham_networks,sato2022pose_cnn_rotation_distribution}. \citet{gilitschenski2020binghamloss} present a new Bingham loss function for orientation distribution learning and \citet{okorn2020learning_orientation_distributions} propose two methods to quantify the uncertainty of orientations for non-symmetric and symmetric objects, respectively. The first method uses an isotropic Bingham distribution to model orientation distribution while the latter learns a multi-modal non-parametric distribution.
\citet{deng2022bingham_networks} propose Deep Bingham Networks as \ac{uq} framework by considering a family of pose hypotheses.
\citet{sato2022pose_cnn_rotation_distribution} present a simple way how a prediction head estimating the parameters of a Bingham distribution can be incorporated into PoseCNN~\citep{xiang2018pose_cnn}.
\citet{manhardt2019pose_ambiguity} use the distribution of pose hypotheses to handle object ambiguities, a goal that was also of interest in \citet{deng2021consecutive_frames} where the orientation distribution is considered while tracking object poses in video frames. In turn, \citet{jeon2023uncertainty_keypoint_selection} use the object ambiguities to estimate confidences for keypoint selection. Further approaches for object pose estimation leverage keypoint confidences to improve the performance and to provide a measure of reliability of the pose estimates \citep{peng2019pnp,huang2022keypoint_confidence,yang2023statistical_guarantees}. 
Others estimate confidences for the pose hypotheses to increase the accuracy of the final average pose result \citep{hu2019segmetation_driven_pose,thalhammer2023cope}.
Recent works use non-parametric distributions to implicitly model the pose distribution in $\mathrm{SE(3)}$. \citet{haugaard2023spyro_pose} present an efficient way to learn pose distributions at different resolution levels. Recently, \citet{zhou20233dnel} combine \mbox{SurfEmb} with inverse graphics and provide a log-likelihood scoring for the estimated poses.
In context of the \ac{uq} methods mentioned in Section \ref{subsec:uq}, these approaches for object pose distribution estimation can be considered as single deterministic approaches to pose \ac{uq}. In contrast to sampling-based \ac{uq} methods, single deterministic approaches do not require multiple forward passes at inference time. However, they are sensitive to the underlying network architecture, training procedure, and training data \citep{gawlikowski2022survey_uncertainty}. Ensemble methods like deep ensembles have been shown to be more robust under datashift and outperform other methods like Monte-Carlo Dropout~\citep{ovadia2019dataschift_uq_comparison}. Furthermore, not all mentioned object pose estimation approaches leveraging uncertainties in their methodology offer uncertainties of the final pose results \citet{uncertainty_driven_pose_estimation}. Also, the quality of the uncertainty estimates is often not explicitly evaluated.
In addition, the incorporation of uncertainties mostly comes with complex changes in established pose estimation methodologies. In contrast, deep ensembles offer a simple approach to \ac{uq}. In this paper, we show how a deep-learning-based object pose estimation approach is extended to additionally quantify uncertainties with deep ensembles.

\section{Background of SurfEmb and Deep Ensembles}\label{sec:background}
Given a multi-stage 6D object pose estimation approach, we evaluate the applicability of the \ac{uq} method of deep ensembles to the task of deep 6D object pose estimation. 
First, in Section \ref{subsec:surfemb}, an introduction to the components of the \mbox{SurfEmb}~\citep{haugaard2022surfemb} method is given, which is chosen as the exemplary method for the task of 6D object pose estimation.
Section \ref{subsec:deepensemble} explains the prerequisites that have to be fulfilled by the ensemble baseline models in order to ensure the creation of a well-calibrated deep ensemble.

\subsection{SurfEmb}\label{subsec:surfemb}
We conduct our experiments using \mbox{SurfEmb}, which is in the top ten of the best performing methods in the BOP challenge 2022~\citep{sundermeyer2023bop}. Like many 6D object pose estimation methods, \mbox{SurfEmb} is a multi-stage approach, which is why the insights gained in this work can be transferred to similar multi-stage approaches as well. It uses the 2D object instance detections produced by \mbox{CosyPose}~\citep{labbe2020cosypose} with Mask\ R-CNN~\citep{he2017mask_rcnn} and trains a deep neural network to predict 2D--3D correspondences that are forwarded to a \ac{pnp} algorithm that estimates the object poses. 
More specifically, \mbox{SurfEmb} learns dense and continuous 2D--3D correspondence distributions by using high-dimensional embeddings of the object surface coordinates. The correspondence network is trained in a self-supervised fashion using a contrastive loss. The positive and negative training examples are provided by the so-called key model, a sinusoidal representation network (SIREN) MLP~\citep{sitzmann2020siren} that transforms a 3D object surface coordinate into a 12D embedding space. The correspondence network or the so-called query model with a U-Net~\citep{ronneberger2015unet} architecture and a ResNet18~\citep{he2016resnet18} backbone is then trained to predict pixel-wise 12D surface embeddings from an input image crop of an object instance. The 2D--3D correspondences are then used in AP3P~\citep{ke2017ap3p}, an algebraic P3P algorithm, to obtain object pose hypotheses followed by a pose hypotheses scoring. Pose hypotheses that exceed a score threshold are locally refined and can be further refined with depth data.

\subsection{Deep Ensembles}\label{subsec:deepensemble}
\citet{deep_ensembles} offer a recipe with three prerequisites that baseline models must fullfill in order to obtain a well-calibrated deep ensemble of models that predicts accurate uncertainties. The prerequisites concern i) the model weights initialization scheme, ii) the scoring rule used during model training, and iii) whether a form of adversarial training is applied.

\textbf{Model Weights Initialization Scheme.} The weights of each model in the ensemble are initialized randomly. The randomness causes the models to reach different modes of the loss during training and thus the ensemble can better cover the posterior distribution of predictions. This is one of the main reasons why deep ensembles work well in practice \citep{fort2020deep_ensembles}.

\textbf{Scoring Rule.} For network training, a probabilistic scoring rule, \mbox{i.e.} a scoring rule that quantifies the quality of the predicted probability distributions, is required. In general, for classifiers that use loss functions maximizing a likelihood, like softmax cross entropy, this is already satisfied. In regression tasks, a Gaussian \ac{nll} can be used.

\textbf{Adversarial Training.} Deep Ensembles incorporate adversarial training~\citep{goodfellow2015adversarial_training} for predictive distribution smoothing. 
During each training step, negative examples are generated based on the current input, which are included in the calculation of the loss. Adversarial Training is proposed as an optional step during model training \citep{deep_ensembles}.

Each individual model in the ensemble must fulfill the first two prerequisites.
Let $N$ be the ensemble size, \mbox{i.e.} the number of ensemble members that are trained in accordance with the above prerequisites and used to produce the ensemble results at inference time. 
In \citet{ovadia2019dataschift_uq_comparison}, it was shown that an ensemble size of $N = 5$ produces good results. Nevertheless, $N$ is an empirical value and should be determined for each application specifically.
The higher the ensemble size, the better the underlying posterior predictive distribution of the ensemble outputs can be approximated.

\section{Methodology}\label{sec:methodology}
In this section, we describe how we apply deep ensembles to \mbox{SurfEmb} and how we evaluate it. 
In Section \ref{subsec:surfembensemble}, the prerequisites described in Section \ref{subsec:deepensemble} are checked and applied to \mbox{SurfEmb}. The evaluation methodology is described in Section \ref{subsec:ensembleevaluation}.

\subsection{SurfEmb Deep Ensemble}\label{subsec:surfembensemble}
In the following, we explain how the three prerequisites described in Section \ref{subsec:deepensemble} are taken into account when using an ensemble of \mbox{SurfEmb}. Next to these prerequisites, the ensemble size $N$ must be defined empirically for our application, which is discussed in the experiments in Section \ref{sec:experiments}.

\textbf{Model Weights Initialization Scheme.} In the original publication of \mbox{SurfEmb}, the ResNet18 backbone of the U-Net architecture of the query model is initialized with pre-trained weights on ImageNet~\citep{deng2009imagenet}. Instead, according to the deep ensemble recipe of Section~\ref{subsec:deepensemble}, we randomly initialize each ensemble query model with different weights drawn from a normal distribution scheme according to \citet{he2015random_init}.
As mentioned in Section \ref{subsec:surfemb}, the key model that provides the targets during training is based on a SIREN MLP. Due to the sensitivity of the used sine non-linearities, the SIREN MLP requires the specification of lower and upper bounds of a uniform distribution from which the weights are randomly drawn.
By randomly initializing the key models in the ensemble, the models learn different realizations of the 12D embedding space of correspondence distributions. 

\textbf{Scoring Rule.} The query model of \mbox{SurfEmb} is trained with a combined loss for the predicted visible object mask and the surface embedding. The object mask is scored by the binary cross entropy while multi-class cross entropy is used as a scoring rule for the surface embeddings. As cross entropy can be considered a proper scoring rule, no modifications are required. 

\textbf{Adversarial Training.} \mbox{SurfEmb} does not incorporate an adversarial training. However, it is trained in a self-supervised manor with a contrastive loss taking both negative and positive examples into account. This training regime has a similar effect on the ensemble results as adversarial training. Therefore, and because this prerequisite is optional, we do not perform any further predictive distribution smoothing.

The resulting \mbox{SurfEmb} ensemble consists of $N = 10$ independent query models of whom each model generates a query, and, based on that, produces an object pose estimate at inference time, forming the ensemble pose estimates.
One major advantage of an ensemble approach for \ac{uq} is that no assumptions are made about the underlying distribution of predictions. In case of object pose estimation, this especially presents an advantage over \ac{uq} methods that predict the parameters of distributions explicitly. As a drawback, one has to overcome the challenging endeavor of extracting meaningful, application specific pose uncertainties.
In case of object symmetries, it is not guaranteed that all pose estimates of the ensemble members refer to the same object symmetry axis. For this purpose, we select the pose in the ensemble prediction with the highest score as reference and align all $N-1$ other poses to that reference based on the known symmetric transformations of the object model.

\subsection{Ensemble Evaluation}\label{subsec:ensembleevaluation}
The pose estimator ensemble is evaluated on the test set of the corresponding training dataset. Let the test dataset $D$ be defined as $D = \left\{ \left( x_t, y_t \right) \right\}_{t=1}^{T}$ where $x_t$ is the $t$-th input data point, \mbox{i.e.} the image crop of an object instance, and $y_t$ the corresponding annotated ground truth pose of the depicted object instance in the camera coordinate frame, composed of a 3D rotation matrix $\mathbf{R}$ and a translation vector $\mathbf{t}$. For the $t$-th entry in the test dataset, the $n$-th ensemble member $H_n$, with $n = \left\{ 1, 2, ..., N \right\}$, outputs a prediction, resulting in a sample of $N$ predictions that are drawn from the posterior predictive distribution. An approximate distribution can be fit to this predictive distribution to get ensemble results that consist of the parameters of the approximate distribution.
In the simplest case, a Gaussian distribution is defined by the mean $\mu_t$ and the standard deviation $\sigma_t$ of the ensemble predictions on the $t$-th dataset entry. Despite the assumption made about the underlying posterior predictive distribution, the standard deviation offers the advantage of being easy to interpret.

\subsection{Uncertainty Evaluation}\label{subsec:quantativeevaluation}
For a consistent evaluation of the ensemble's means and standard deviations, we compute reliability diagrams or calibration plots \citep{deep_ensembles,kuleshov2018confidence_plot}.
The reliability diagram for an ensemble or any forecaster $H$ that predicts a cumulative density function (CDF) $F_t$ of the $t$-th dataset entry is computed based on the following assumption: $H$ is well calibrated on dataset $D$ if the predicted CDFs match the empirical CDFs when the dataset size $T$ tends towards infinity \citep{kuleshov2018confidence_plot}. Given $M$ chosen expected confidence levels $0 \leq p_1 < p_2 < ... < p_M \leq 1$, the corresponding observed confidence level $\hat{p}_j$ to each threshold $p_j$ is calculated by computing the empirical frequency \citep{kuleshov2018confidence_plot}:
\begin{equation}\label{eq:confidencelevels}
    \hat{p}_j = \frac{\left| \left\{ y_t | F_t \left( y_t\right) \leq p_j, t = 1,...,T \right\} \right|}{T}
    \enspace .
\end{equation}
If $H$ is well calibrated, the values $\left\{ \left( p_j, \hat{p}_j \right)\right\}_{j=1}^M$ form a straight line which passes through the origin and has slope $1$ \citep{kuleshov2018confidence_plot}.
In our case, where we estimate the probability density function in terms of the mean and standard deviation of the ensemble predictions on the $t$-th dataset entry, we first compute the corresponding predicted CDF $F_t$ and get the empirical CDF by drawing the $t$-th target value $y_t$ from $F_t$.

Based on this reliability diagram, we propose an uncertainty evaluation metric that takes into account the area between the perfect calibration where the target and predictive distributions match and the actually observed confidence levels. We call the metric the \ac{uscore}. Whereas in case of a perfect uncertainty calibration the area is zero, the worst case uncertainty calibration corresponds to the possible maximum value of this area which is $A_{max} = 0.25$. 
Therefore, based on these lower and upper bounds, we propose to quantify the calibration quality by
\begin{equation}\label{eq:uscore}
    \mathrm{\ac{uscore}} = 1 - A / A_{max}
    \enspace ,
\end{equation}
where $A$ is the area estimated from the reliability diagram and calculated by using the composite trapezoidal rule:
\begin{equation}\label{eq:trapez}
    A = \int_0^1 f \left( p \right) dp \approx \frac{\Delta p}{2} \sum_{j=1}^{M} \left(  f \left( p_{j-1} \right) +  f \left( p_{j} \right) \right)
    \enspace ,
\end{equation}
where $f \left( p \right) = \left| \hat{p} - p \right|$ is the absolute difference between the observed confidence level $\hat{p}_j$ estimated using Equation \eqref{eq:confidencelevels} and the expected confidence level $p_j$. The computation increment is set to $\Delta p = 0.1$.  
\ac{uscore} is bound between $\left[0, 1 \right]$, where a higher value indicates a better calibration.
Consequently, the metric is easy to interpret and facilitates a comparison or even a ranking of the different methods.
The estimated area $A$ can be interpreted as the calibration error and is similar to the \ac{ece}~\citep{guo2017calibration_moderna}, a popular network calibration error metric for semantic segmentation and classification tasks. Both, the \ac{ece} and $A$ in Equation \eqref{eq:uscore} are computed based on the differences between the observed accuracy or confidence and the predicted or expected confidence. In the sense that the \ac{ece} depends on the number of bins, \ac{uscore} also depends on the chosen computation increment $\Delta p$ in the calculation of the observed confidence levels of the reliability diagram in Equation \eqref{eq:confidencelevels}. On the other hand, this dependency implies flexibility with respect to the choice of the assumed underlying distribution. In combination with sampling-based \ac{uq} methods, \ac{uscore} can be used to find the optimal sample size. Note that in contrast to the \ac{ece}, \ac{uscore} can be applied to any calibration plot, regardless whether the underlying task is regression, classification, or segmentation.
The parameterization of the cumulative distribution, which is assumed for the calculation of the observed confidence levels, can be freely chosen, since in the case of deep ensembles the choice is not restricted. Therefore, varying distributions for the orientation and position component can be considered in the computation of \ac{uscore} for a deep ensemble of an object pose estimator.
Here, we use Gaussians to parameterize the distributions of both the orientation and position.

\begin{figure}[t]
    \begin{center}
   \includegraphics[width=0.8\linewidth]{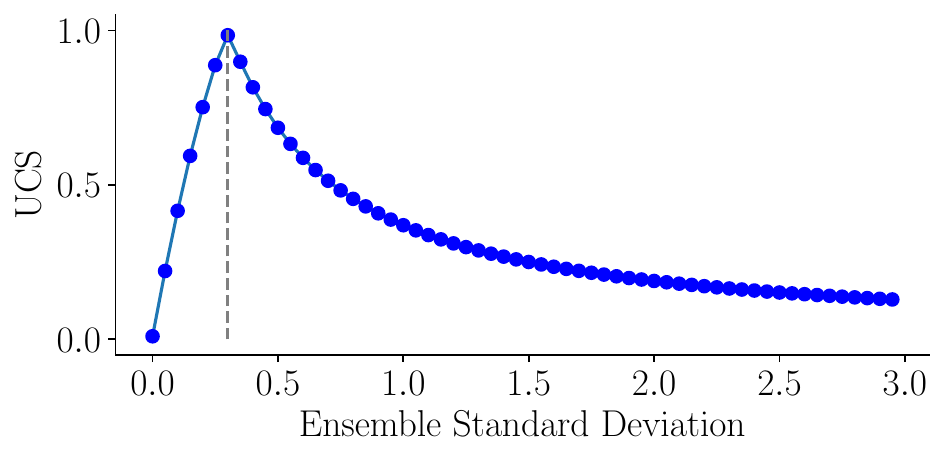}
    \end{center}
   \caption{\ac{uscore} for simulated uncertainty predictions with a ground truth standard deviation $\sigma_{true} = 0.3$ (dashed line). For a perfect calibration, where the simulated and predicted uncertainty match, UCS is close to $1$.}
\label{fig:ucstest}
\end{figure}

Since the ground truth posterior uncertainty of the ensemble outputs is unknown, we use synthetic data to evaluate the performance of our proposed quality score \ac{uscore}. 
Given a dataset size of $T = 10000$, we sample each $t$-th target from a uniform distribution. The corresponding output distributions are represented by a mean and the standard deviation $\sigma_{pred}$. For each target, the ensemble output mean is sampled from a Gaussian distribution centered on the target and with a standard deviation of $\sigma_{true} = 0.3$. Based on the reliability diagram, we expect that $\mathrm{\ac{uscore}} = 100.0\ \%$, if $\sigma_{true} = \sigma_{pred}$ for all $T$ dataset entries. The more $\sigma_{pred}$ differs from the expected value of $\sigma_{true}$, the more should \ac{uscore} decrease. The results of the simulation are shown in Figure \ref{fig:ucstest} and confirm our expectations.
In case that the predicted uncertainty matches the ground truth standard deviation, the $\ac{uscore}$ is $0.990$. For wrong predictions the \ac{uscore} is smaller. As the standard deviations do not have an upper bound, \ac{uscore} slowly converges to zero if the estimated uncertainties are too large. In case of a few outliers ($0.01\Unit{\%}$ of $T$) that successfully target the ground truth but where $\sigma_{pred} \neq \sigma_{true}$, \ac{uscore} is not affected significantly and therefore proves to be robust to outliers.

\section{Experiments}\label{sec:experiments}
\begin{figure}[t]
    \centering
    \subfloat[Scene 1]{%
        \includegraphics[width=0.48\linewidth]{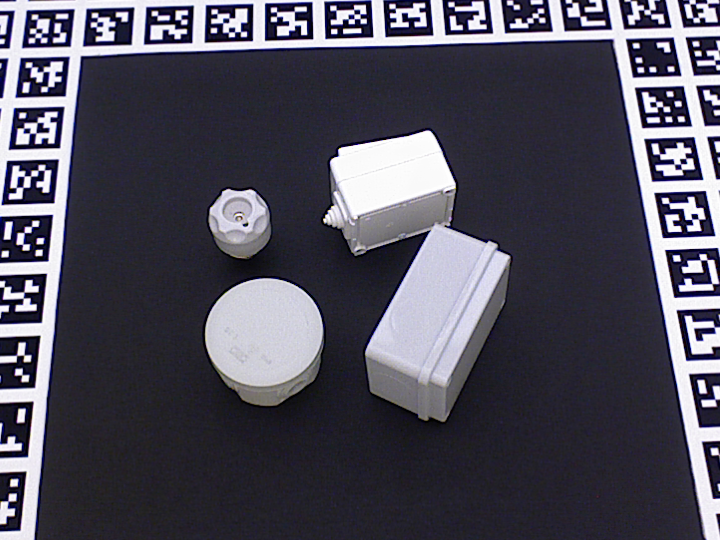}%
        \label{fig:tlessexamplescene1}%
    }%
    \hfill%
    \subfloat[Scene 18]{%
        \includegraphics[width=0.48\linewidth]{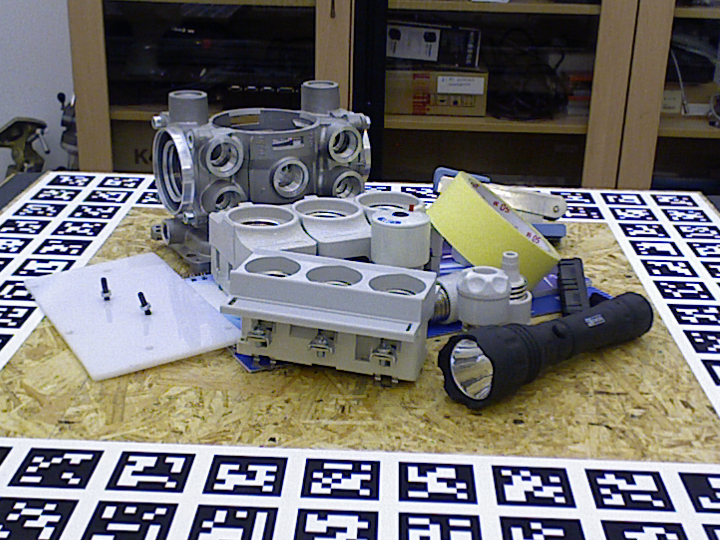}%
        \label{fig:tlessexamplescene18}%
    }%
    \caption{Two examples of RGB images from different scenes of the \ac{bop} test dataset of T-LESS.}
\label{fig:tlessexamples}
\end{figure}

We conducted experiments on the T-LESS~\citep{hodan2017tless} and YCB-V~\citep{xiang2018pose_cnn} datasets. Both are part of the \ac{bop} challenge and include photo\-realistic rendered training images of randomly sampled cluttered scenes that were added to the datasets as part of the \ac{bop} challenge 2020 \citep{hodan2020bop_challenge} and are used to train the models. While T-LESS consists of 30 industrial parts that are largely textureless and in many cases symmetric, the YCB-V dataset contains 21 objects of daily life. Both datasets include CAD object models. The \ac{bop} test dataset of T-LESS is composed of 20 cluttered scenes, for each of which 50 real test images are provided. Two example RGB images are shown in Figure \ref{fig:tlessexamples}. The \ac{bop} test images of YCB-V are sampled from 12 of the 92 video scenes of the original dataset. Because not all annotated ground truth targets are visible, we only consider target objects where at least 10\Unit{\%} of the object instance surface is visible and the visible object instance part is represented by at least 1024 pixels in the image. This results in 6423 and 4121 valid ground truth samples for T-LESS and YCB-V, respectively. As these criteria are also used during model training, this test data subset can be considered as in-domain. In contrast, test data points of object instances that are visible by less than 10\Unit{\%} and whose visible surface masks have fewer than $1024$ pixels form an out-of-domain test dataset.
In Section \ref{subsec:evalpose}, we evaluate the quality of the pose estimates of the trained ensemble members for T-LESS and YCB-V on their respective \ac{bop} test datasets. In Section \ref{subsec:evaluncertainty}, the estimated ensemble distributions are analyzed.

\subsection{Evaluation of the Ensemble Pose Estimates}\label{subsec:evalpose}
\begin{figure*}[!t]
\begin{floatrow}\CenterFloatBoxes
\floatbox{table}[\FBwidth]{\caption{Evaluation results of the differently trained \mbox{SurfEmb} models on the BOP test datasets of T-LESS and YCB-V, both without (RGB) and with depth refinement (RGB-D). Shown are the reproduced $AR$ of the models trained and provided by the authors of \mbox{SurfEmb} (Pretrained), the mean $AR$ of the randomly initialized ensemble members (Baseline), and the evaluation results of the mean poses of the ensembles (Ensemble).}%
\label{tab:bopresults}}
        {\resizebox{!}{1.25cm}{%
            \bgroup
            \def\arraystretch{1.15}
            \begin{tabular}{ll|ccc|ccc}
                \hline
                 & & \multicolumn{3}{c}{RGB} & \multicolumn{3}{|c}{RGB-D} \\
                \hline
                Dataset & Scores & Pretrained & Baseline & Ensemble & Pretrained & Baseline & Ensemble \\
                \hline
                \multirow{4}{*}{T-LESS}
                & $\mathrm{AR}_{MSPD}$ $\uparrow$ & $0.846$ & $0.844 \pm 0.002$ & $\mathbf{0.851}$ & $0.851$ & $0.848 \pm 0.002$ &  $\mathbf{0.855}$\\
                & $\mathrm{AR}_{MSSD}$$\uparrow$ & $0.554$ & $0.556 \pm 0.010$ & $\mathbf{0.586}$ & $0.821$ & $0.815 \pm 0.003$ &  $\mathbf{0.823}$\\
                & $\mathrm{AR}_{VSD}$ $\uparrow$ & $0.525$ & $0.527 \pm 0.010$ & $\mathbf{0.561}$ & $0.784$ & $0.779 \pm 0.002$ &  $\mathbf{0.790}$\\
                & $\mathrm{AR}$ $\uparrow$ & $0.642$ & $0.642 \pm 0.007$ & $\mathbf{0.666}$ & $0.818$ & $0.814 \pm 0.002$ &  $\mathbf{0.823}$\\
                \hline
                \multirow{4}{*}{YCB-V}
                & $\mathrm{AR}_{MSPD}$ $\uparrow$ & $0.759$ & $0.750 \pm 0.002$ & $\mathbf{0.764}$ & $0.789$ & $0.780 \pm 0.003$ &  $\mathbf{0.794}$\\
                & $\mathrm{AR}_{MSSD}$$\uparrow$ & $\mathbf{0.498}$ & $0.483 \pm 0.008$ & $0.489$ & $0.850$ & $0.844 \pm 0.003$ &  $\mathbf{0.852}$\\
                & $\mathrm{AR}_{VSD}$ $\uparrow$ & $\mathbf{0.433}$ & $0.417 \pm 0.008$ & $0.421$ & $0.757$ & $0.748 \pm 0.003$ &  $\mathbf{0.762}$\\
                & $\mathrm{AR}$ $\uparrow$ & $\mathbf{0.563}$ & $0.550 \pm 0.005$ & $0.558$ & $0.799$ & $0.791 \pm 0.003$ &  $\mathbf{0.803}$\\
                \hline
            \end{tabular}
            \egroup
        }
}
\ffigbox[\FBwidth]
       {\includegraphics[width=0.65\linewidth]{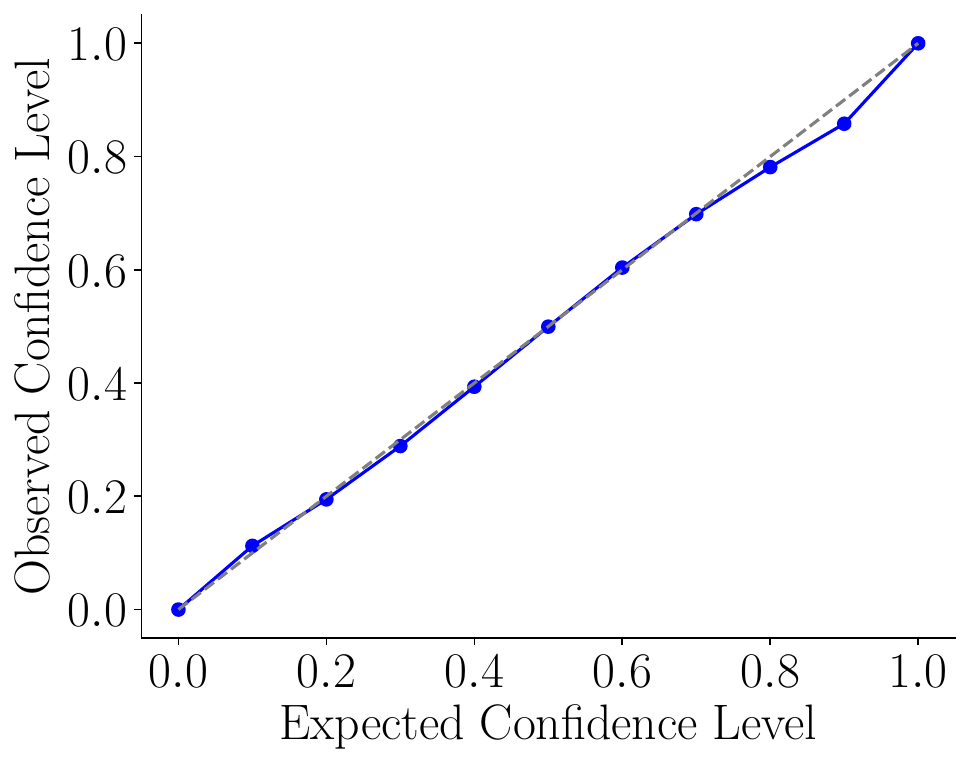}}{%
  \caption{Reliability diagram of the T-LESS query model ensemble with the optimal ensemble size of eight ensemble members. The perfect calibration as the diagonal is represented by the dashed gray line.}%
  \label{fig:queryconfidence}%
}
\end{floatrow}
\end{figure*}

To ensure that the overall quality of the pose estimates of the trained ensembles is not affected by the random initialization, we evaluated each of the ensemble members, which we call the baseline models, separately. We also evaluated the pose average as the mean over all ensemble members. For evaluation, we applied the \ac{bop} error metrics $MSPD$, $MSSD$, and $VSD$ \cite{hodan2020bop_challenge}. 
In Table \ref{tab:bopresults}, the results with and without depth refinement are compared to the scores of the reproduced \mbox{SurfEmb} model that is provided by the authors, the mean of the scores achieved by the ensemble baseline models, and the scores for the estimated mean poses of the ensemble are reported. 
Both ensembles trained on T-LESS and YCB-V consist of ten ensemble members each.
The scores are defined as the average recall ($AR$) of the \ac{bop} error metrics $MSPD$, $MSSD$, and $VSD$. While the $MSPD$ error measures the perceivable discrepancy and, therefore, is relevant for augmented reality applications, the $MSSD$ error measures the maximum pose error in the 3D space and is especially relevant for robot manipulation \citep{hodan2020bop_challenge}. Both metrics take object symmetries into account. The $VSD$ is the visual surface discrepancy \cite{hodan2020bop_challenge}. Surprisingly, the poses of the randomly initialized ensemble members achieve similar scores as the ones estimated by the provided \mbox{SurfEmb} models for T-LESS and YCB-V that were initialized with pretrained weights on ImageNet. It seems that in this case, pretraining does not improve the quality of the predictions. Furthermore, it can be observed that ensembling the poses slightly improves the quality of the prediction, a phenomenon that is often taken advantage of in knowledge distillation, where the knowledge of an ensemble is compressed into a single model to overcome the ensemble drawback of high computational costs \cite{hinton2015distilling}.
\thisfloatsetup{floatwidth=\paperwidth,rowfill=yes,margins=hangboth}

\subsection{Evaluation of the Ensemble Uncertainty}\label{subsec:evaluncertainty}
\begin{figure}[t]
    \centering
    \subfloat[Ensemble orientations with and without local refinement]{%
        \includegraphics[width=0.5\linewidth]{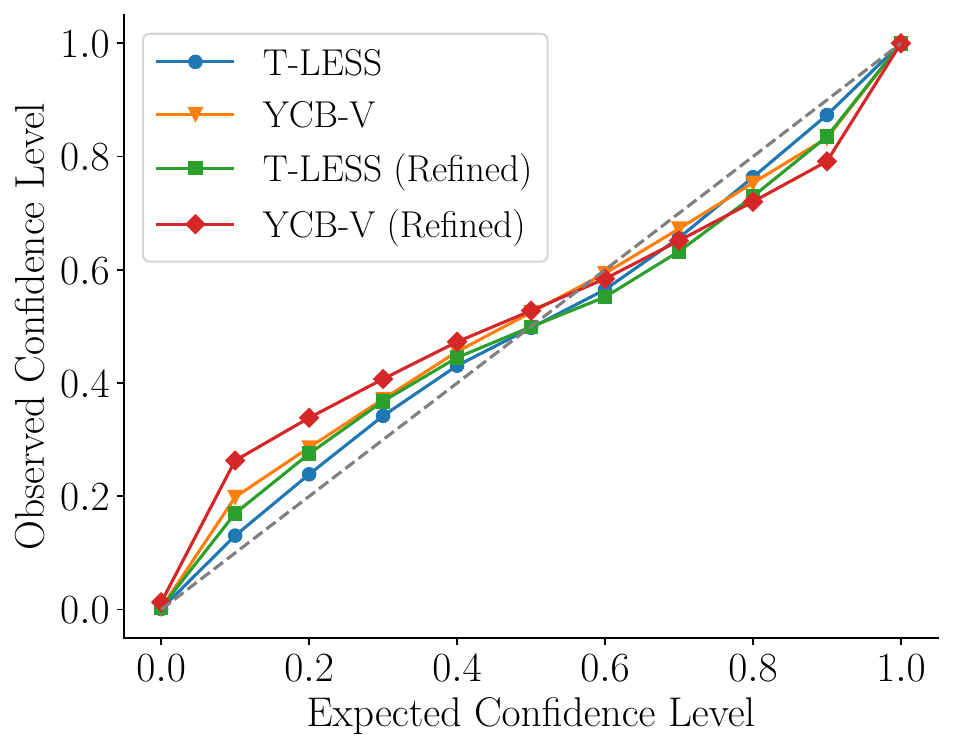}%
        \label{fig:orientationconfidence}%
    }%
    \hfill%
    \subfloat[Locally refined ensemble positions with and without depth]{%
        \includegraphics[width=0.5\linewidth]{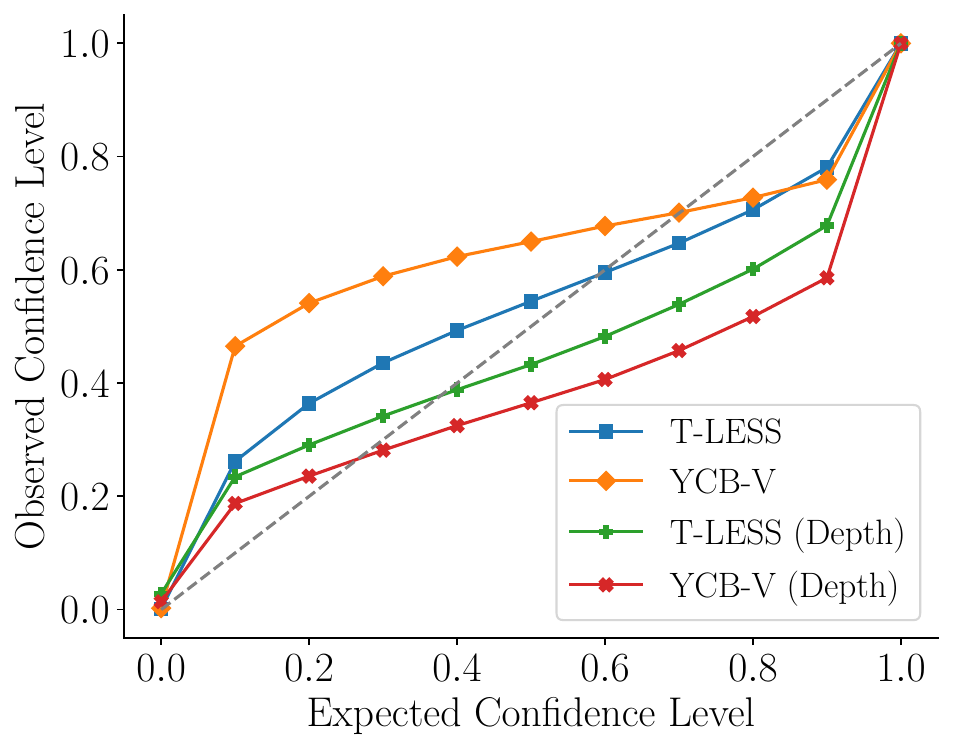}%
        \label{fig:positionconfidence}%
    }%
    \caption{Reliability diagrams of the estimated ensemble orientation and position components on T-LESS and YCB-V. The perfect calibration is represented by the dashed gray line.}
\label{fig:reliabilitydiagramsrefinements}
\end{figure}

To eliminate possible influences of the 2D object detection on the evaluation of the pose estimation ensemble, the evaluation of the ensemble results is done on object instance image crops based on the ground truth. 
In Figure \ref{fig:queryconfidence}, the reliability diagram described in Section \ref{sec:methodology} is shown for the T-LESS ensemble query model outputs, meaning the 12D embedded dense correspondence distributions. 
The reliability diagram can be interpreted as follows: For instance, with an expected confidence level of $0.10$, we observe an actual confidence level of $0.11$, meaning that for $11\ \%$ of the T-LESS test data points the CDFs of the Gaussian distributions output a probability of $\leq 10\ \%$. The Gaussian distributions are parameterized by the ensemble means and standard deviations and evaluated on the targets.
The observed confidences of the ensemble queries are estimated based on the pixel-wise mean and the standard deviation of each embedding dimension. As the targets produced by the key models and the predicted queries are not in the same range, they were normalized with their minimum and maximum values, respectively. The ensemble outputs are predicted on the ground truth object instance crops of the $6423$ visible samples of the T-LESS test dataset. As the area between the plotted curve of the observed confidence levels and the diagonal that represents a perfect calibration is very small, the query model seems to be very well calibrated. Accordingly, it has a high corresponding calibration score of $\mathrm{\ac{uscore}} = 96.0\ \%$. It can be observed that for low expected confidence levels the observed confidence is slightly larger than the expected value. Analogously, for high expected confidence levels, the observed confidence level is slightly lower than expected.
Based on the predicted queries by the ensemble members on the ground truth object instance image crops, the pose results of each ensemble member are computed. 
Figure \ref{fig:reliabilitydiagramsrefinements} shows the reliability diagrams of the estimated orientation in form of rotation matrices and position components on T-LESS and YCB-V. It can be observed that, overall, the T-LESS ensemble seems to be better calibrated than the YCB-V ensemble.
In Figure \ref{fig:orientationconfidence} shows that the quality of the calibration of the orientation component decreases with the local refinement step, both on T-LESS and YCB-V. The unrefined poses achieve a \ac{uscore} of $88.5\ \%$ and $80.6\ \%$ while the refined estimates score $79.6\ \%$ and $69.3\ \%$ on T-LESS and YCB-V, respectively.
In contrast, the position component is only lightly affected by the local refinement and decreases the \ac{uscore} of the unrefined estimates by $3.1\ \%$ on T-LESS and by $0.8\ \%$ on YCB-V. While the depth refinement does not influence the orientation estimates, it improves the quality of the estimated position component that also affects its calibration, as it is shown in Figure \ref{fig:positionconfidence}.
While the \ac{uscore} of the position component on YCB-V increases from $33.6\ \%$ to $44.3\ \%$ when they are refined with the depth data, the \ac{uscore} on T-LESS decreases from $65.2\ \%$ to $57.6\ \%$. The reason behind this will be part of future work.

\begin{figure}[t]
    \centering
    \subfloat[Different orientation representations on T-LESS]{%
        \includegraphics[width=0.5\linewidth]{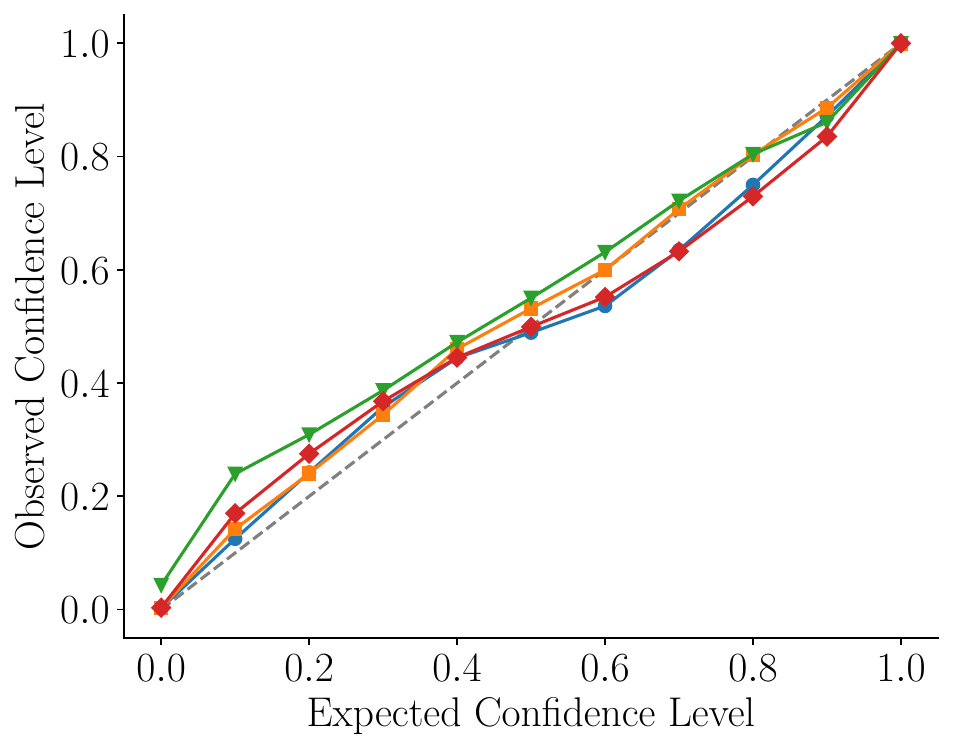}%
        \label{fig:tlessconfidencerepresentations}%
     }%
     \hfill%
     \subfloat[Different orientation representations on YCB-V]{%
        \includegraphics[width=0.5\linewidth]{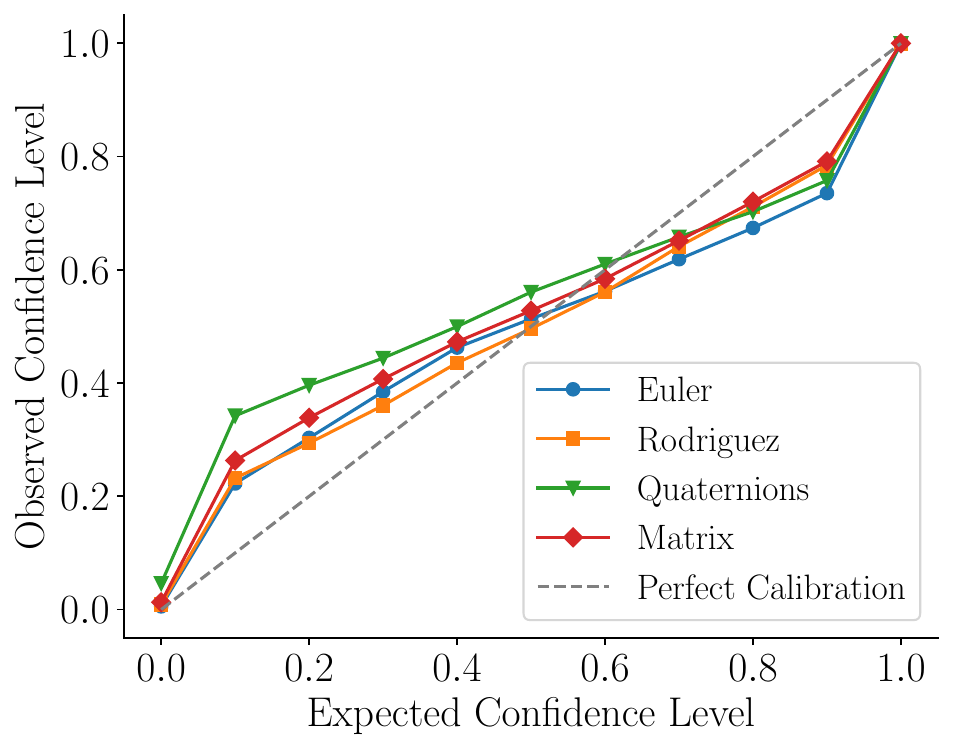}%
        \label{fig:ycbvconfidencerepresentations}%
     }
    \caption{Reliability diagrams of the locally refined ensemble orientation components on T-LESS and YCB-V.}
\label{fig:reliabilitydiagramsrepresentations}
\end{figure}

In Figures \ref{fig:tlessconfidencerepresentations} and \ref{fig:ycbvconfidencerepresentations}, the reliability diagrams of different orientation representations of the pose estimates on the ground truth image crops of T-LESS and YCB-V are shown. The poses are unrefined so that the influence of the orientation representations can be better observed and other influences are as much reduced as possible. The four chosen representations are quaternions, Euler angles, Rodriguez' axis-angle representation, and the rotation matrix. These representations were selected based on their importance in orientation and pose estimation tasks and their interpretability. 
Out of the four representations, Rodriguez' axis-angle representation achieves the highest \ac{uscore} with $90.5\ \%$ and $77.1\ \%$ on T-LESS and YCB-V, while the quaternion representation scores the lowest with $76.7\ \%$ and $59.3\ \%$.
The calibration of the position component is noticeably decreased in comparison to the orientation component, regardless of the representation. 

\section{Discussion}
The reliability diagrams of the T-LESS query model ensemble in Figure \ref{fig:queryconfidence} and of the orientation components on both datasets in Figure \ref{fig:reliabilitydiagramsrefinements} show that the deep ensembles are well calibrated. This is also reflected in the high values for UCS. The almost perfect calibration of the query model ensembles is reduced during the follow-up steps of the pose estimation pipeline of \mbox{SurfEmb}. 
This leads to the conclusion that, while deep ensembles are easy to apply in general, this notion may not be transferred to an ensemble of a pose estimator with multiple stages and a combination of deep learning and algorithms. It may be circumvented by using an end-to-end trainable pose estimator where the \ac{pnp} algorithm is implemented as part of the trainable network architecture as it was done in GDRNet~\citep{wang2021gdr_net} or by using error propagation.
However, Figure \ref{fig:queryconfidence} shows that the ensemble results of the query model on the T-LESS dataset are well calibrated, demonstrating that in this case for \ac{uq} with deep ensembles the stage consisting of the 2D object detector does not need to be included. It has to be noted that the position components of the pose estimates is detrimental to the overall calibration.
In the reliability diagrams of different orientation representations on T-LESS and YCB-V, shown in Figures \ref{fig:tlessconfidencerepresentations} and \ref{fig:ycbvconfidencerepresentations}, it can be seen that the choice of the representation has an influence on the imperfections and thus the calibration. Remarkable is the decrease of the calibration quality in case of a representation as quaternions, which may be due to the fact that the assumed normal distribution with one standard deviation value per element is not sufficient.

\section{Conclusion}
In this work, we applied the state-of-the-art deep learning \ac{uq} method of deep ensembles to \mbox{SurfEmb}, one of the top-per\-forming multi-stage deep 6D object pose estimation approaches, and evaluated the result on the T-LESS dataset.
The adaptation of \mbox{SurfEmb}'s correspondence network to the deep ensemble methodology is straightforward and we find that the ensemble on T-LESS is very well calibrated. However, the following \ac{pnp} implementation, pose refinement strategies, and pose representations reduce the quality of the estimated predictive uncertainty. Furthermore, we introduced UCS, a novel metric to quantify the estimated uncertainty for regression tasks. UCS is easy to interpret and facilitates a comparison or even a ranking of the different methods.
In future work, we want to extend the experiments to other pose estimation methods. Also, we want to investigate the influence of the error propagation of the network ensemble predictions through the P$n$P(-RANSAC) stage of the pose estimation pipeline. This may be done by using a differentiable \ac{pnp} implementation like EPro-P$n$P~\citep{chen2022epro_pnp} or a deep learning variant like Patch-P$n$P~\citep{wang2021gdr_net}.

{
	\begin{spacing}{1} 
		\normalsize
		\bibliography{references} 
	\end{spacing}
}

\end{document}